\documentclass[letterpaper]{article} 
\usepackage{aaai2026}  
\usepackage{times}  
\usepackage{helvet}  
\usepackage{courier}  
\usepackage[hyphens]{url}  
\usepackage{graphicx} 
\usepackage{natbib}  
\usepackage{caption} 
\usepackage{algorithm}
\usepackage{algorithmic}

\usepackage{newfloat}
\usepackage{listings}
\DeclareCaptionStyle{ruled}{labelfont=normalfont,labelsep=colon,strut=off} 
\floatstyle{ruled}
\newfloat{listing}{tb}{lst}{}
\floatname{listing}{Listing}
\usepackage[utf8]{inputenc} 
\usepackage[T1]{fontenc}    
\usepackage{hyperref}       
\usepackage{url}            
\usepackage{booktabs}       
\usepackage{amsfonts}       
\usepackage{nicefrac}       
\usepackage{microtype}      
\usepackage{xcolor}         

\usepackage{hyperref}
\usepackage{bm}
\usepackage{listings}
\usepackage[most]{tcolorbox}
\usepackage{longtable}
\usepackage{multirow}

\usepackage{amsmath}
\usepackage{amssymb}
\usepackage{mathtools}
\usepackage{amsthm}

\usepackage{tabularx}
\usepackage{array}
\usepackage{enumitem}

\usepackage{etoc}             
\usepackage[title,titletoc]{appendix} 

\usepackage[textsize=tiny]{todonotes}
\usepackage{prettyref}
\newcommand{\pref}{\prettyref}

\newrefformat{fig}{Fig.~\ref{#1}}
\newrefformat{tab}{Table~\ref{#1}}
\newrefformat{sec}{Section~\ref{#1}}
\newrefformat{app}{Appendix~\ref{#1}}
\newrefformat{alg}{Algorithm~\ref{#1}}
\newrefformat{property}{Property~\ref{#1}}
\newrefformat{theorem}{Theorem~\ref{#1}}
\newrefformat{lemma}{Lemma~\ref{#1}}
\newrefformat{corollary}{Corollary~\ref{#1}}
\newrefformat{proposition}{Proposition~\ref{#1}}
\newrefformat{def}{Definition~\ref{#1}}
\newrefformat{eq}{equation~(\ref{#1})}

\def\our{\texttt{FAITH}}

\definecolor{mybackground}{RGB}{241, 238, 252}
\definecolor{note}{RGB}{238, 248, 248}
\definecolor{noteblue}{RGB}{234, 240, 251}
\definecolor{refblue}{RGB}{3, 20, 110}
\definecolor{maroon}{rgb}{0.76, 0.13, 0.28}

\newcolumntype{L}[1]{>{\raggedright\arraybackslash}p{#1}} 

\usepackage{bibentry}

\title{Assessing Automated Fact-Checking for Medical LLM Responses\\with Knowledge Graphs}

\author{
 Shasha Zhou\textsuperscript{\rm 1,2}, Mingyu Huang\textsuperscript{\rm 1}, Jack Cole\textsuperscript{\rm 1}, Charles Britton\textsuperscript{\rm 2}, Ming Yin\textsuperscript{\rm 3}, Jan Wolber\textsuperscript{\rm 4}, Ke Li\textsuperscript{\rm 1}\\
}
\affiliations{
  \textsuperscript{\rm 1} Department of Computer Science and Engineering, University of Electronic Science and Technology of China\\
  \textsuperscript{\rm 2} Department of Computer Science, University of Exeter \\
  \textsuperscript{\rm 3} College of Medicine and Health, University of Exeter \\
  \textsuperscript{\rm 4} Department of Computer Science, Purdue University \\
  \textsuperscript{\rm 5} GE HealthCare\\
  \{sz484, m.huang, jc1417, k.li\}@exeter.ac.uk, charles.britton@nhs.scot, mingyin@purdue.edu, jan.wolber@gehealthcare.com
}
\begin{document}

\maketitle

\begin{abstract}
    The recent proliferation of large language models (LLMs) holds the potential to revolutionize healthcare, with strong capabilities in diverse medical tasks. Yet, deploying LLMs in high-stakes healthcare settings requires rigorous verification and validation to understand any potential harm. This paper investigates the reliability and viability of using medical knowledge graphs (KGs) for the automated factuality evaluation of LLM-generated responses. To ground this investigation, we introduce \our, a framework designed to systematically probe the strengths and limitations of this KG-based approach. \our\ operates without reference answers by decomposing responses into atomic claims, linking them to a medical KG, and scoring them based on evidence paths. Experiments on diverse medical tasks with human subjective evaluations demonstrate that KG-grounded evaluation achieves considerably higher correlations with clinician judgments and can effectively distinguish LLMs with varying capabilities. It is also robust to textual variances. The inherent explainability of its scoring can further help users understand and mitigate the limitations of current LLMs. We conclude that while limitations exist, leveraging KGs is a prominent direction for automated factuality assessment in healthcare.
\end{abstract}

\section{Introduction}
\label{sec:intro}
Large language models (LLMs) have emerged as powerful tools with potential to revolutionize healthcare. These models demonstrate strong capabilities across diverse medical tasks~\cite{SinghalSMW23,Google24,Nori23,VanVBLJ24,Liu2025,McDuffST+25}. However, deploying LLMs in the high-stakes healthcare domain, where precise and trustworthy information is critical, calls for rigorous evaluation. Due to inherent characteristics of LLMs like the vast output scale, automated evaluation methods have become increasingly necessary. These methods complement traditional validation approaches like clinical studies~\cite{SinghalSMW23,WangLX+23}, which are often too slow or limited in scope to keep pace with rapidly evolving LLM technologies. Foremost among these evaluation concerns is ensuring the factual integrity of LLM-generated medical content~\cite{WangLY+23,ThirunavukarasuTEH23,Lee23,Shen23,AugensteinBC+24}. This is crucial as LLMs can produce plausible yet dangerously inaccurate information, known as hallucinations~\cite{Farquhar24,JiLFYSXIBMF23}. Such inaccuracies undermine stakeholder's trust and pose significant barriers to clinical adoption and regulatory approval of LLM applications.

\begin{figure*}
    \centering
    \includegraphics[width=.92\linewidth]{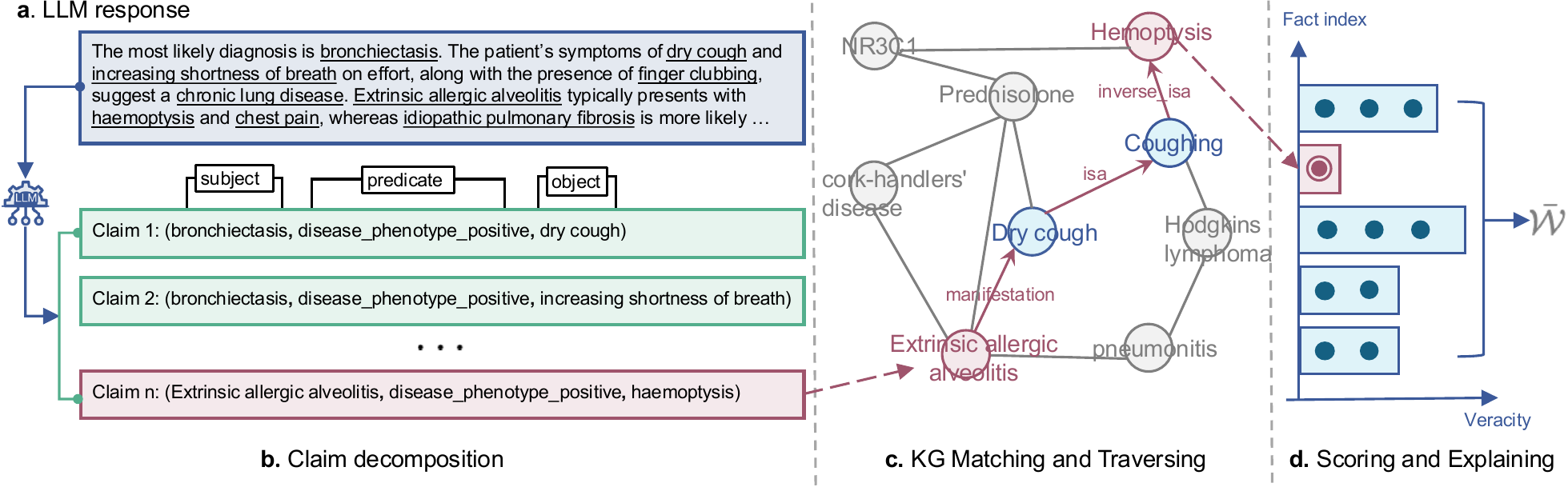}
    \caption{\small\textbf{Overview of \our.} \textbf{(a)} An example medical content generated by an LLM. \textbf{(b)} \our\ processes structured claims relating different medical entities in the response, which are automatically extracted by an LLM. \textbf{(c)} Extracted claims are matched with nodes (maroon) in a medical KG. Paths (maroon) and intermediate nodes (blue) linking the entities are identified. \textbf{(d)} A factuality score for each claim is computed based on the path characteristics and edge semantics. The response score is aggregated from individual claim scores.}
    \label{fig:demo}
\end{figure*}

Traditionally, automated evaluation of LLM outputs has relied on natural language processing (NLP) metrics such as BLEU~\cite{PapineniRWZ02} and BERTScore~\cite{ZhangKWWA20}. Yet, the usefulness of these metrics in specialized domains like healthcare is limited as they require manually crafted reference responses that are rarely available in practice. In addition, it has been reported in various recent studies that they often correlate poorly with clinician judgments~\cite{SinghalSMW23,VanVBLJ24,Liu2025}, probably because these metrics focus on \textit{overall} lexical or semantic similarities with reference texts. In contrast, human experts prioritize evaluating the factual accuracy of \textit{specific} claims within LLM-generated responses (e.g., the statement ``dry cough is a symptom of bronchiectasis", \pref{fig:demo}a-b) stylistic variations or peripheral details.

Alternative approaches for assessing factuality involve using other LLMs or supervised models as evaluators~\cite{ChernCCYFZHGNL23,MinKLLYKIZH23,WangRGAMMPABNG23}. However, relying on LLMs for factuality evaluation can also be problematic, as they are susceptible to data-poisoning attacks that compromise their reliability~\cite{Alber2025}, and these evaluators themselves can suffer from hallucinations. Supervised models detect factual errors by identifying patterns from historical training data, assuming future data will follow the same patterns. However, this assumption often fails in unforeseen error types, reducing their generalizability.

In recent years, various medical knowledge graphs (KGs) have been developed to represent diverse validated medical knowledge as collections of nodes (entities) and edges (relationships)~\cite{ChenSSWLHWWY25, PoulainB24, HamzaAALK25}. They have been successfully applied to assist the generation of medical responses via retrieval-augmented generation (RAG)~\cite{GaoXG+23}. Inspired by this, in this paper we hypothesize that these same medical KGs may also be helpful in assisting the fact-checking of LLM-generated medical contents. Yet, unlike generation tasks, fact-checking with KGs can be more complex as it requires multi-hop exploration through complex relational paths within the KG. Additionally, effective fact-checking must quantify how well extracted claims align with structured medical knowledge, making simple lookups inadequate.

To systematically assess the potential use of KG in medical fact-checking for LLMs, we developed \our\ (\textbf{\underline{f}}act-\textbf{\underline{a}}ware evaluat\textbf{\underline{i}}on of LLM-genera\textbf{\underline{t}}ed contents in \textbf{\underline{h}}ealthcare), an unsupervised, reference-free framework based on medical KGs. \our\ first decomposes an LLM-generated response into atomic medical claims (\pref{fig:demo}b). It then uses entity resolution techniques~\cite{ChristophidesEP21} to map the medical entities within these claims to their corresponding nodes in a medical KG (\pref{fig:demo}c). Next, \our\ traverses the KG to identify the \textit{evidence paths} connecting the linked entities, based on which the factuality of each claim will be assessed. Finally, an overall factuality score for the entire response is aggregated from the individual claim scores, while preserving per-claim scores for nuanced interpretability (\pref{fig:demo}c-d).

We evaluated the effectiveness of \our\ and various existing fact-checking approaches on four established medical question-answering (QA) tasks with both quantitative and subjective evaluations. We also investigated whether \our\ is readily usable to improve the safety of LLM-based medical systems by employing \our\ as a thresholding mechanism. We additionally assessed \our's broader utility by applying it to medical summarization and fact verification tasks and conducted ablation analysis on its core components. With these experiments, we observed several features of this KG-based fact-checker compared to conventional ones:

\begin{itemize}
    \setlength{\itemsep}{0pt}
    \setlength{\parskip}{0.3pt}
    \item It operates without reference answers or supervised training, making it highly suitable for real-world settings.
    \item It correlates significantly better with clinician judgments.
    \item It is robust to noise in the input texts, and can effectively distinguish LLMs of varying capabilities.
    \item It offers better interpretability than existing approaches by pinpointing which particular claim is susceptible.
    \item Yet, the reliability of \our\ places high demands on the KG quality, which is expected to improve as newer medical KGs come out in the future.
\end{itemize}

\section{Related Work}
\label{sec:related}
\subsection{Factuality Evaluation of LLMs in Medicine}

Methods for evaluating the factuality of LLM-generated medical contents generally fall into two categories.

The first line of work compares LLM outputs to reference answers using NLP metrics. Examples include BLEU~\cite{PapineniRWZ02} and ROUGE~\cite{Lin04} for lexical overlap, and BERTScore~\cite{ZhangKWWA20} for contextual similarity in embedding space. To further incorporate domain-specific knowledge into evaluation, MEDCON~\cite{YimFA+23} uses QuickUMLS~\cite{SoldainiG16}—a tool for extracting biomedical concepts—to allow for a more informed evaluation. imapScore~\cite{WangZ0024} adopts the similar idea, but instead uses an LLM for medical concept extraction. However, in practice, human references are hardly available, the application of these metrics is therefore largely limited. In contrast, \our, is reference-free and much more flexible.

Another line of work employs supervised machine learning methods to detect factual errors. Earlier ones like FED~\cite{MehriE20} and DEB~\cite{SaiMAK20} adopted BERT-like architectures. Yet, these approaches hardly generalize beyond their original training data. Later works used more advanced GPT-series models for evaluation, e.g., GPTScore~\cite{FuNJ024}, G-Eval~\cite{LiuIXWXZ23}, and GPT-judge~\cite{LinHE22B}. More recently, LLM-Eval~\cite{LinC23} adopts a multi-aspect evaluation schema for more comprehensive evaluation. Specifically in the medical domain, \citeauthor{VladikaHM25} leverage multi-turn interactions with LLMs to reach a binary decision on a single claim. However, these LLM judges can also suffer from hallucinations. In addition, all these model-based approaches lack sufficient explainability to the evaluation. In contrast, \our\ is unsupervised, grounded on medical facts, and offers inherent explainability to pinpoint the errors.

Also related are those works regarding hallucination detection of LLMs~\cite{JiLFYSXIBMF23, ZhangLC+23}, which is a broader topic. Our work specifically targets one type of hallucination that stems from factual errors, and complements research on other types, e.g., confabulations~\cite{Farquhar24} or instruction misalignment~\cite{ZengYG0G024}. Combining \our\ with other hallucination-detection methods can enable more systematic safety oversight.

\subsection{Knowledge Graph-Based Fact-Checking}
In general domains, external knowledge base like KGs, have been widely used for fact-checking~\cite{Gad-Elrab0UW19, CiampagliaSR+15,ShiralkarFMC17, FiondaP18, ShiW16, KimPKJTC23, DammuNDKRCS24, HuangZCL25, yu2025plantscience, CIM}. Early approaches assessed the ``quality'' of paths linking entities within each claim. For instance, KL~\cite{CiampagliaSR+15} suggests that paths involving generic entities provide weaker support and uses node degrees for evaluation. KL-REL~\cite{ShiralkarFMC17} extends this by considering the semantic similarity of edge labels in the path to the predicate in the original claim. Other methods like TransE~\cite{BordesUGWY13} employ global representations of entities and relationships to better capture hierarchical structures in KGs. While established, these methods often assume that a single fact is being checked each time and that its entities are directly available in the KG, making them not directly applicable to LLMs' free-text outputs.

\section{The FAITH Evaluation Framework}
\label{sec:methods}

\subsection{Preliminaries}

\noindent \textbf{Claims of fact.} A \textit{claim of fact} is a triplet $t=(s,r,o)$, where $s$, $o$ denote the subject and object, respectively, and $r$ is the predicate (i.e., relation between $s$ and $o$). For example, the triplet $(\text{Barack Obama}, \textit{was\_born\_in}, \text{Hawaii})$ expresses the claim that Barack Obama was born in Hawaii.

\noindent \textbf{Knowledge graphs (KGs).} A \textit{knowledge graph} $\mathcal{G}=(\mathcal{E}, \mathcal{R})$ is composed of a set of entities (nodes) $\mathcal{E}$ and relations (edges) $\mathcal{R}$ among them. It can be regarded as a collection of factual claims defined as $\mathcal{T} = \mathcal{E} \times \mathcal{R} \times \mathcal{E}$.

\noindent \textbf{Knowledge paths.} A \textit{knowledge path} $p$ between entity $e_1$ and $e_k$ is a sequence of alternating entities and relations, denoted as $p=(e_1, r_1, e_2, r_2, \ldots, r_{k-1}, e_k)$, where $e_i \in \mathcal{E} (1 \leq i \leq k)$, $r_i \in \mathcal{R} (1 \leq i < k)$, and $(e_{i}, r_{i}, e_{i+1}) \in \mathcal{T}$ for $1 \leq i < k$. This path consists of $k$ entities and $k-1$ relations. The path length is defined as the number of relations it contains, $k-1$.

\noindent \textbf{Computational fact checking.} Given an LLM-generated response $\mathcal{D}$ containing a set of medical factual claims $\{t_i\}_{i=1}^{n}$ and a KG $\mathcal{G}$, the goal of \textit{computational fact checking} is to determine the factuality of each claim triplet $t_i=(s_i, r_i, o_i)$ in $\mathcal{D}$ by leveraging the knowledge paths $p_i$ in $\mathcal{G}$.

\subsection{Key Modules of FAITH}

As shown in \pref{fig:demo}, \our\ comprises 4 key modules:

\noindent \textbf{Medical claim extraction.} To extract medical claims $\{t_i\}_{i=1}^{n}$ from an LLM response $\mathcal{D}$, we use GPT-4o for this purpose due to its strong extraction capability~\cite{DagdelenLW24,PolakM24}. Specifically, we use a multi-phase prompting strategy (Appendix B.1)\footnote{Appendix is available at: https://zenodo.org/records/17603819} that first identifies all medical entities in $\mathcal{D}$ and then determines relationships among them. To enhance performance, we employ in-context learning with $5$ clinician-crafted examples. We also adopt a multi-round prompting~\cite{KuratovBA+24,GraphRAG24} to increase recall, whilst a redundancy-inducing prompt is used to let the model critically reanalyze the extracted claims with suspicion to mitigate hallucination~\cite{PolakM24}.

\noindent \textbf{Medical entity matching.} In classic fact-checking literature~\cite{CiampagliaSR+15,ShiralkarFMC17}, benchmarking datasets are often generated from the KG used in the algorithm. Thus, entities in this data can always match with KG nodes. Yet, in real-world applications, entities in the LLM responses often cannot directly match with nodes in the KG (e.g., `haemoptysis' and `Hemoptysis' in~\pref{fig:demo}c). To resolve these discrepancies, we use entity resolution techniques~\cite{OlivierR22} via the Unified Medical Language System (UMLS)~\cite{Bodenreider04}. UMLS is a comprehensive ontology linking synonymous medical terms from over $200$ different biomedical vocabulary sources via concept unique identifiers (CUIs). We use the UMLS API to map entities from both claims and nodes in $\mathcal{G}$ to standardized CUIs. This provides a strong foundation of reliable, consensual knowledge, which is essential for a fact-checking system. Claims with unmatched entities are labeled as `unverifiable' and excluded from further evaluation. This represents a conservative design choice, ensuring that \our\ only scores claims for which it has a high-confidence knowledge base.

\noindent \textbf{KG traversal and factual evaluation.} After entity matching, we traverse the KG to extract the knowledge paths $\mathcal{P}= \{p_i\}_{i=1}^n$ connecting the resulting nodes, where each path $p$ is a sequence $(e_1, r_1, e_2, r_2, \ldots, r_{k-1}, e_k)$. Given the vast scale and dense connectivity of medical KGs, an exhaustive search of all possible paths between two entities is computationally intractable. Therefore, we constrain our search to the shortest path, which represents the most direct and salient connection. In addition to computational feasibility, the shortest paths are also considered optimal for maximizing the information content of the connection~\cite{CiampagliaSR+15}. The obtained path $p_i$ can then serve as evidence to evaluate the factuality of the corresponding medical claim $t_i$.

It is often considered that longer paths indicate weaker factuality~\cite{CiampagliaSR+15, ShiralkarFMC17}, but path length alone is insufficient. This is because the semantics of relations between entities can significantly affect the factuality of the claim. For example, the relations `\textit{indication}' and `\textit{contraindication}' have totally different meanings when directly linking two entities. To include this, we introduce a relation semantic similarity $\mathcal{S}(p,t)$ to assess the congruence between the semantics of the knowledge path $p$ and the medical claim $t$. We use the mean cosine similarity between the embeddings of the relations in $p$ and the relation in $t$:
\begin{equation}
    \mathcal{S}(p, t) = \frac{1}{k-1} \times \sum_{i=1}^{k-1} cos\_sim(r_i, \hat{r}),
\end{equation}
where $k-1$ is the length of $p$, $r_i$ represents the $i$-th relation in $p$, and $\hat{r}$ is the predicate in $t$.

However, this is still inadequate to fully capture the factuality of the medical claim $t$. \citet{CiampagliaSR+15} suggest that paths involving generic entities provide weaker support to a claim. Moreover, \citet{ShiralkarFMC17} emphasize the importance of considering the co-occurrence of each relationship $r_i$ in $p$ with $t$'s predicate to enhance the accuracy of factuality assessment. Building on these insights, \our\ incorporates the entity centrality (via PageRank, $PR$~\cite{BrinP98}) and relationship co-occurrence ($u(r_i, \hat{r})$) derived from contracted line graphs. Putting these together, the factuality score for a given medical claim $t$ related to a knowledge path $p$ is given by:
\begin{equation}
    \label{eq:score}
    \mathcal{W}(p,t)=\mathcal{S}(p,t)\Bigg[\sum_{i=2}^{k-1}\frac{e^{\alpha PR(e_i)}}{u(r_{i-1}, \hat{r})}+\frac{1}{u(r_{k-1}, \hat{r})}\Bigg]^{-1},
\end{equation}
where $e_i$ and $r_i$ are the $i$-th entity and relation in $p$, respectively, and $\alpha$ is a scaling constant set to $100$ in this study (see Appendix B.2). The term $u(r_i,\hat{r})$ is calculated as the co-occurrence between two relations based on contracted line graph, with higher values indicating tighter relations.

\noindent \textbf{Scoring and interpretation.} Finally, the overall \our\ score for the whole LLM-generated response $\mathcal{D}$, which contains the set of claims $\{t_i\}_{i=1}^n$ is aggregated from the individual factuality scores for each claim involved:
\begin{equation}
    \bar{\mathcal{W}}(\mathcal{P}, \mathcal{D}) = \frac{1}{n}\times \sum_{p_i \in \mathcal{P}, t_i\in \mathcal{D}}\mathcal{W}(p_i,t_i).
\end{equation} 
\our\ generates factuality scores in the range of $[-1, 1]$, where positive values indicate alignment with established medical knowledge while negative values suggest contradictions. The absolute value reflects how strongly entities in a claim are interconnected. Our scoring system ensures interpretability by explicitly linking each assessment to the KG paths examined (see examples in Appendix D.1).

\section{Experimental Setup}
\label{sec:setup}

\noindent \textbf{Baseline methods.} We compare \our\ against the following baselines: $\blacktriangleright$ \textbf{NLP metrics:} This includes BLEU-4~\cite{PapineniRWZ02} and ROUGE-L~\cite{Lin04}. We also included BERTScore~\cite{ZhangKWWA20} which leverages contextual embedding model (here \href{https://www.voyageai.com/}{voyage-3-large}, a top-performer on MTEB leaderboard~\cite{MuennighoffTMR23}) to evaluate the semantic similarity between reference and input. Additionally, we considered MEDCON~\cite{YimFA+23} as it is specifically designed for medical scenarios. $\blacktriangleright$ \textbf{Computational fact-checking methods:} These include $3$ established methods: KL~\cite{CiampagliaSR+15} and KL-REL~\cite{ShiralkarFMC17}, and TransE~\cite{BordesUGWY13}. $\blacktriangleright$ \textbf{LLM-based method:} We consider FActScore~\cite{MinKLLYKIZH23} and imapScore~\cite{WangZ0024} to represent this category. Note that as KL, KL-REL, and TransE cannot be directly applied to free texts, we preprocess their inputs with our entity matching module. NLP metrics and imapScore are provided with reference responses for evaluation.

\noindent \textbf{LLMs.} To compare the efficacy of our proposed approach with the baselines, we considered $5$ representative LLMs. This includes $2$ established proprietary models, GPT-4o-mini and GPT-4o~\cite{Hurst24}. We also included two open-source models, Llama 3-8B and Llama 3.1-8B~\cite{DubeyJP+24}. We additionally included a specialized model, \href{https://huggingface.co/aaditya/Llama3-OpenBioLLM-8B}{OpenBioLLM}, which is fine-tuned on medical texts and specifically designed to answer medical questions in an informative way. 

\noindent \textbf{Medical tasks.} To assess \our's efficacy, we focused on medical QA tasks as they purely demand a high factual accuracy on medical knowledge. Other tasks, such as clinical decision making, further require advanced reasoning capabilities. While these are useful for evaluating the overall medical capabilities of LLMs, they can introduce additional confounding factors to our evaluation, which focuses on factuality. In this spirit, we selected $4$ established datasets: $i)$ MedQA~\cite{JinPO+20} $ii)$ MMLU~\cite{HendrycksBBZMSS21}, $iii)$ \href{https://www.medschools.ac.uk/medical-licensing-assessment/preparing-for-the-ms-akt/practice-exam-for-the-ms-akt}{MS-AKT}, and $iv)$ LiveQA~\cite{AbachaAPD17}. These span both multiple-choice (MedQA, MMLU, MS-AKT) and open-ended questions (LiveQA), with varying levels of complexity and coverage. For more details, see Appendix C.3.

\noindent \textbf{Medical KG.} While many medical KGs exist~\cite{LuGZW25,Bodenreider04, Louden20,CheungNS+96}, many of them are tailored to focus on specific medical problems, whereas our fact-checking would desire an encyclopedic KG that covers a wide range of medical information (see discussion in Appendix B.3). To this end, we consider the UMLS~\cite{Bodenreider04} 2025AA version for this paper. It integrates knowledge from over $200$ standard biomedical vocabulary sources, (e.g., \href{https://www.ncbi.nlm.nih.gov/mesh/}{MeSH}, \href{https://digital.nhs.uk/services/terminology-and-classifications/snomed-ct}{SNOMED CT} and \href{https://hpo.jax.org/}{Human Phenotype Ontology}), incorporating more than $23$ million relationships among $3.4$ million concepts.

\section{Experimental Results}
\label{sec:results}

\begin{figure*}[h!]
    \includegraphics[width=\linewidth]{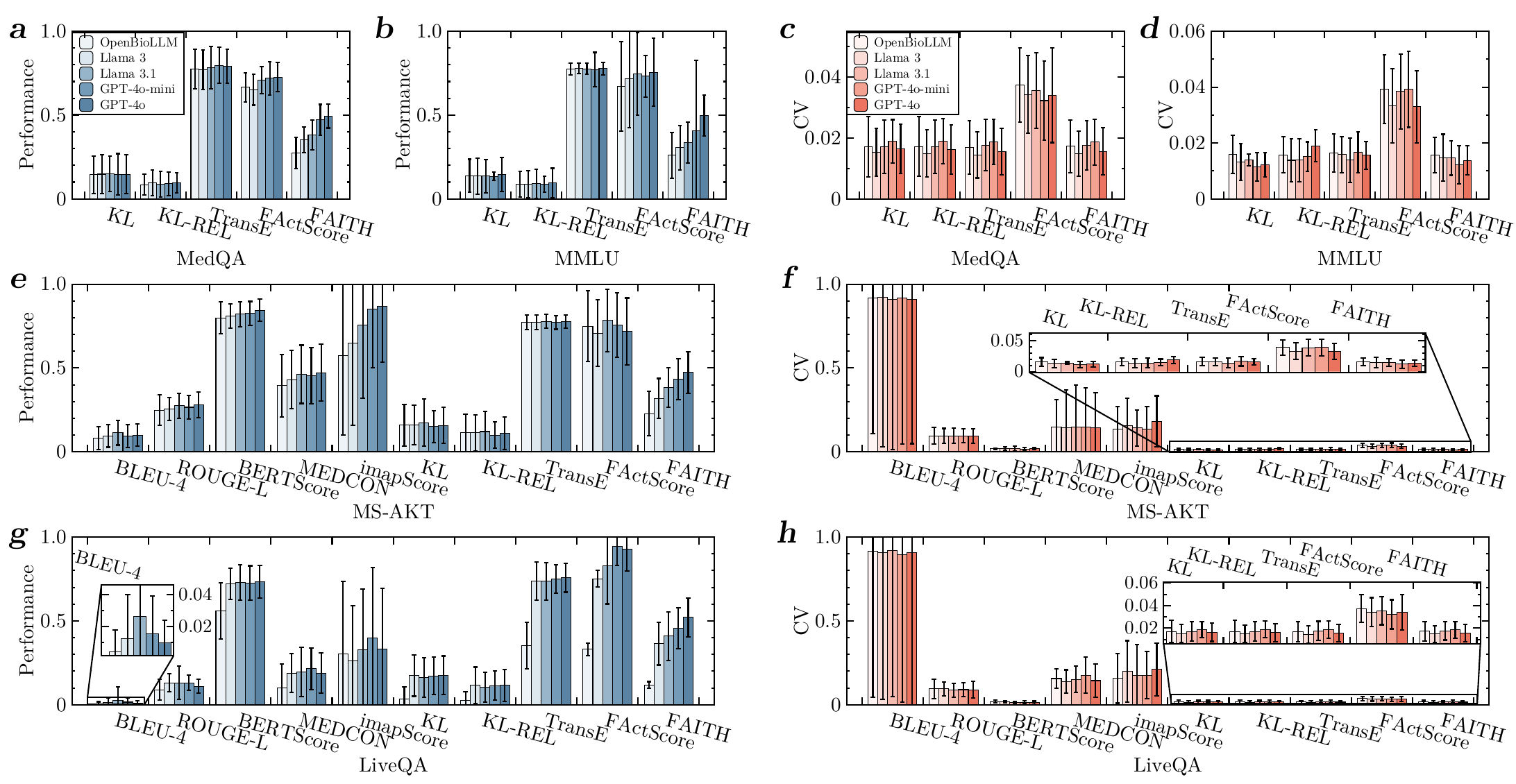}
    \caption{\small\textbf{\our\ effectively distinguishes between LLMs and is robust to noise.} This figure shows the mean factuality scores assigned by \our\ and baseline metrics to responses from five LLMs on four datasets: MedQA \textbf{(a)}, MMLU \textbf{(b)}, MS-AKT \textbf{(e)}, and LiveQA \textbf{(g)}. A reliable metric should assign distinguishable scores across different models. Panels \textbf{c, d, f,} and \textbf{h} display the corresponding coefficients of variation (CV) of these scores under noisy conditions, introduced by generating $10$ paraphrased versions per response.}
    \label{fig:re_1a}
\end{figure*}

\subsection{Reliability and Validity of FAITH}
\label{sec:rq1}
An effective evaluation metric for LLM-generated responses in healthcare should satisfy several key desiderata from practitioners. To be specific, it should: $i)$ clearly distinguish responses produced by LLMs with different levels of capabilities; $ii)$ be robust to the variance or noise in the responses that are not relevant to the evaluation; and $iii)$ align with the consensus of experienced clinicians. We thus evaluate \our\ on these $3$ dimensions.

\noindent \textbf{Distinguishing LLM capabilities.} We first probed \our's ability to differentiate $5$ LLMs with varying medical capabilities~\cite{ZhangSJ+24, BoggavarapuSV+24, KeJE+25, ChanGS+24} by evaluating their responses on $4$ medical QA tasks. We find that metrics such as ROUGE-L, BERTScore, and TransE on average rated all $5$ models similarly (paired $t$-test $p>0.05$;~\pref{fig:re_1a}a, b, e, and g). While other metrics like BLEU-$4$, MEDCON, and FActScore achieved statistically significant differences between some of the models, this discrepancy does not persist across all models and tasks. For example, FActScore fails to distinguish between Llama 3.1 and GPT-4o-mini on MedQA (\pref{fig:re_1a}a).
In contrast, \our\ consistently distinguished responses from all $5$ LLMs ($p<0.004$). This indicates that the approach is sensitive enough to capture meaningful differences in the factual quality of outputs from models with varying capabilities.

\noindent \textbf{Robustness to variance.} We then test \our's robustness to variations in response phrasing. To do this, we used GPT-4o to generate $10$ paraphrased versions for each response from the $5$ LLMs (Appendix C.5). The paraphrasing instructions emphasized modifying sentence structures, pronouns, and prepositions while preserving medical terminology and relationships. An ideal metric should yield low variance across such paraphrased responses. We selected the coefficient of variation (CV) for this purpose, as it offers a measure of relative variability. Across all scenarios, \our\ exhibits consistently low CV (\pref{fig:re_1a}c, d, f, and h), with the average $\pm$ s.d. being $0.014\pm0.005$. In contrast, metrics like BLEU-4 showed much higher variability ($0.910\pm0.862$). This suggests that \our\ is more robust to the variations in phrasing.

\noindent \textbf{Correlation with clinician judgments.} Finally, to determine the alignment of \our\ with human clinicians, we recruited $20$ UK-based clinicians to manually grade the responses generated by LLMs. Each clinician holds a medical degree and possesses at least $5$ years of clinical practice experience. In close consultation with these experts, we identified $3$ key evaluation dimensions: $i)$ factuality, $ii)$ relevance, and $iii)$ potential harm. Each dimension was rated on a $5$-point Likert scale (with $1$ being the lowest and $5$ the highest; Appendix C.6), and an additional overall assessment also used the same scale. Following these criteria, each clinician assessed $5$ responses, which were randomly drawn from a candidate question pool of $16$ questions from MS-AKT dataset. These responses could either originate from Llama 3.1, GPT-4o, or official answers. To ensure rating consistency, each response was evaluated by at least two clinicians, with Cohen's $\kappa=0.64$, indicating substantial inter-rater agreement.

\begin{figure}[t!]
    \centering
    \includegraphics[width=\linewidth]{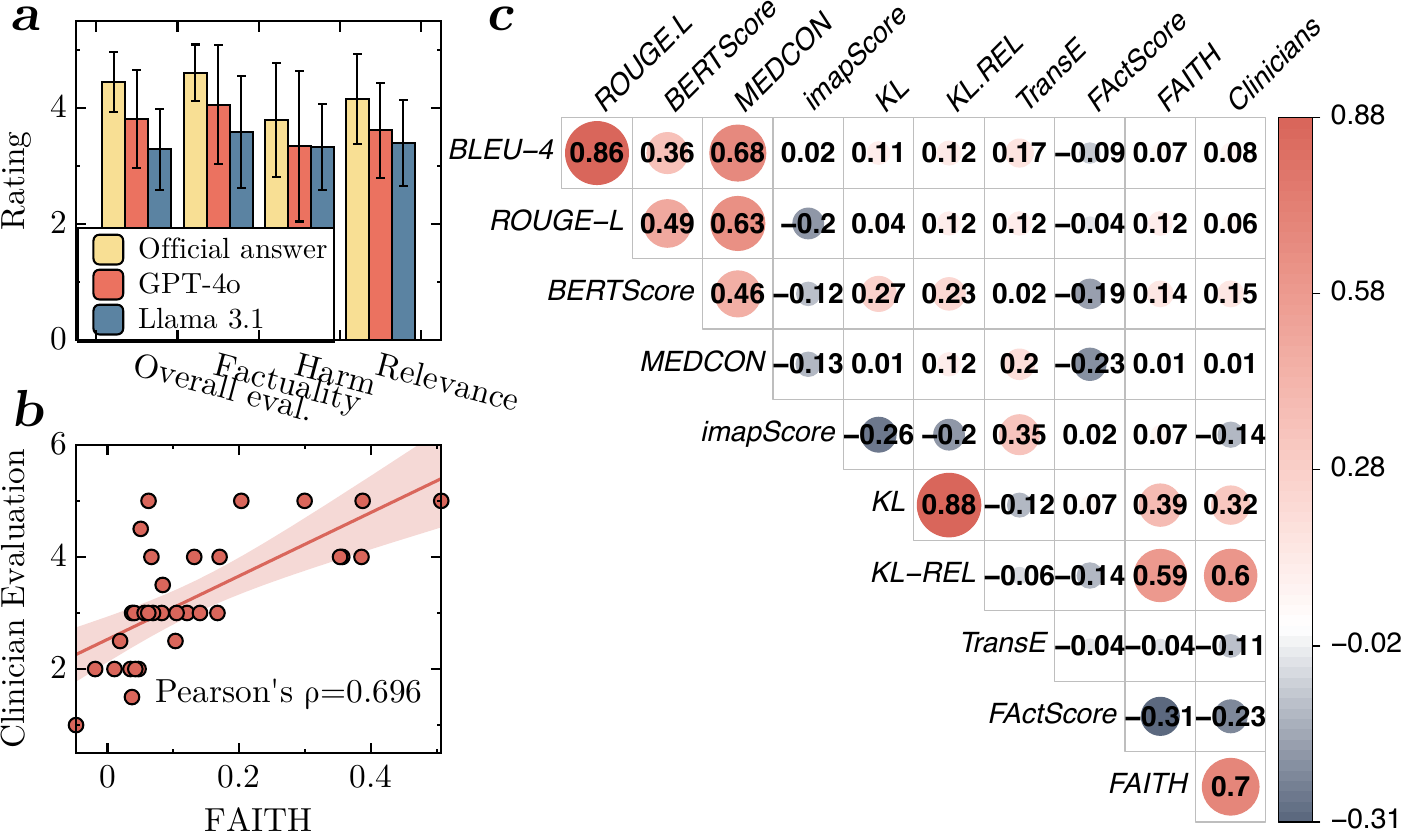}
    \caption{\small\textbf{\our\ exhibits highest correlation with clinician judgments.} \textbf{a,} Clinician evaluation scores for answers to $16$ questions generated by GPT-4o, Llama 3.1, and official answers. \textbf{b,} Scatter plots correlating clinician judgments with scores from \our. Linear regression fits are shown with $95\%$ confidence intervals. \textbf{c,} Pearson correlation coefficients ($\rho$) between clinicians and scores from \our\ and various baselines.} 
    \label{fig:rq_1c}
    \vspace{-2em}
\end{figure}

As shown in~\pref{fig:rq_1c}a, clinician assessments consistently ranked official answers highest across all evaluation axes, followed by GPT-4o, and then Llama 3.1. To quantify alignment between automated metrics and these human judgments, we computed Pearson's correlation coefficient $\rho$. The results show that \our\ scores achieve a considerably higher correlation with clinicians (Pearson's $\rho=0.696$;~\pref{fig:rq_1c}b, c) compared to all baselines. This provides strong evidence for the effectiveness of the KG-based evaluation approach. Notably, baselines like BLEU-4 only has $\rho=0.081$, consistent with previous findings~\cite{VanVBLJ24}.

\subsection{Explainability of FAITH}
\label{sec:rq2}

We evaluate \our's interpretability with $2$ aspects: $i)$: the faithfulness of its explanations to medical consensus, and $ii)$: the utility of these explanations for analyzing LLM behaviors.

\begin{figure}
    \centering
    \includegraphics[width=\linewidth]{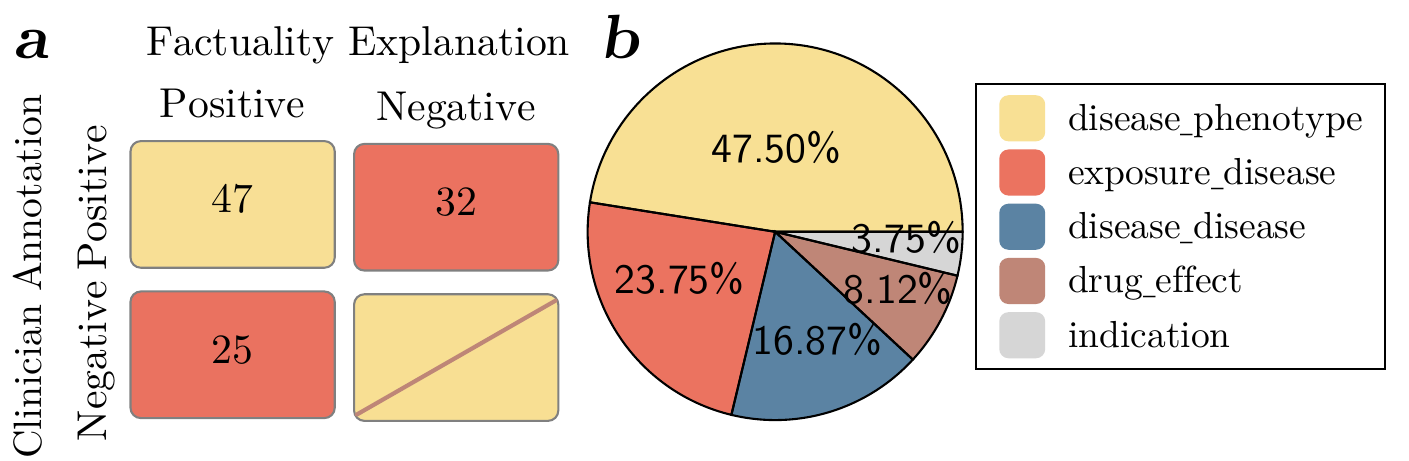}
    \caption{\small\textbf{Explainability of \our\ via faithful error identification and LLM limitation analysis.} \textbf{a,} Alignment between clinician-identified incorrect claims and \our's lowest-scoring claims in LLM responses, shown by a confusion matrix. \textbf{b,} Distribution of the top-5 most frequent KG relation types linked to incorrect claims in GPT-4o's responses, as identified by \our.}
    \label{fig:rq_2}
\end{figure}

\noindent\textbf{Faithfulness to medical consensus.} For the first aspect, we additionally asked clinicians to annotate the one most incorrect medical claim in each justification they had previously rated the factuality score lower than $5$. These annotations were manually converted to fact triplets to be comparable with the factuality explanations generated by \our. We evaluated \our's ability to pinpoint these specific annotations, achieving a precision of $0.65$, recall of $0.59$, and F1 score of $0.62$ (\pref{fig:rq_2}a). It is important to consider that the selection of the most incorrect claim by clinicians can involve subjective considerations (e.g., specific expertise, perceived claim importance). We therefore evaluated its performance in localizing these claims and found that in $83.6\%$ of these justifications, the erroneous statement pinpointed by clinicians was ranked by \our\ among the top-$5$ lowest-scoring. These findings suggest that \our's explanations could be faithful to current medical consensus, offering valuable, granular insights into response inaccuracies.

\noindent\textbf{Utility in analyzing LLM limitations.} Such explanations offered by \our\ can be potentially utilized to offer insights into limitations of LLMs. To show this, we investigated the top-$5$ most frequent edge types in the KG that are associated with the most incorrect claims made in GPT-4o's responses, as identified by \our. As shown in \pref{fig:rq_2}b, nearly half of the identified inaccuracies involve phenotypical features of diseases. Such errors are particularly prominent when LLMs attempt to determine a disease from patient signs, which underscores their current challenges in the rigorous clinical task of differentiating between numerous potential pathologies for a given initial complaint~\cite{HagerJ+24}. Furthermore, frequent errors in establishing causal relationships between diseases and exposures highlight LLM tendencies to overstate associations or assert causality without sufficient evidence~\cite{GriotH+25, YudiC24}. This granular error typology (Appendix D.3), is vital for guiding target LLM refinement in complex medical applications.

\subsection{Practical Utility of FAITH for Safeguarding}
\label{sec:rq3}

\begin{figure}
    \centering
    \includegraphics[width=\linewidth]{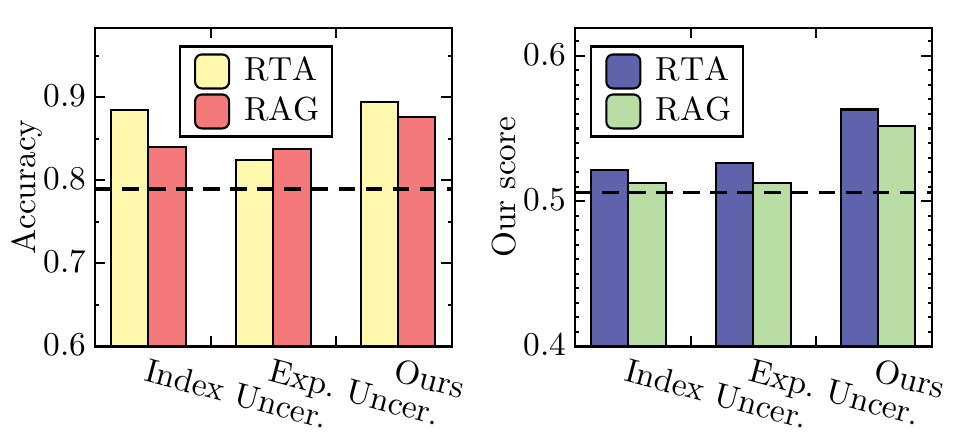}
    \caption{\small\textbf{\our\ enhances LLM factuality via selective intervention.} GPT-4o performance on MedQA using Reject-to-Answer (RTA) or Retrieval-Augmented Generation (RAG). Interventions triggered by either \our\ scores or a model uncertainty baseline. \textbf{a,} Question-answering accuracy. \textbf{b,} \our\ factuality scores for responses. Metrics plotted against the percentage (x-axis) of responses selected for intervention.}
    \label{fig:re_3q}
\end{figure}

Beyond evaluation, \our\ can be used to improve the factuality and safety of LLM-generated medical responses. We show this in two intervention scenarios. In the first one, we use \our\ scores below a set threshold to activate a Reject-to-Answer (RTA) protocol,  filtering out responses with low factuality~\cite{SinghalSMW23, WenYF+24}. In the second one, we augment the LLMs with a RAG module~\cite{GaoXG+23} that retrieves relevant information from the medical KG when potential factual errors are detected. Model uncertainty served as a baseline triggering mechanism (see Appendix B.4). For both scenarios, intervention thresholds were determined by percentiles of the \our\ or uncertainty score distribution in the dataset (ranging from $5\%$ to $50\%$).

Our findings show that by applying the RTA protocol, both the accuracy of answers and the factuality of the justification generated by GPT-4o could be significantly improved (\pref{fig:re_3q}). While uncertainty-based thresholding could also lead to boosted results, the improvements are not as significant as those achieved by \our. Likewise, RAG could also enhance both metrics, particularly the factuality of the justification. Instead of employing RAG for all questions, the use of \our\ as a thresholding mechanism can offer a more cost-effective solution in practice. This advantage also persists if we consider other LLMs (Fig. A16).

\begin{figure}
    \centering
    \includegraphics[width=\linewidth]{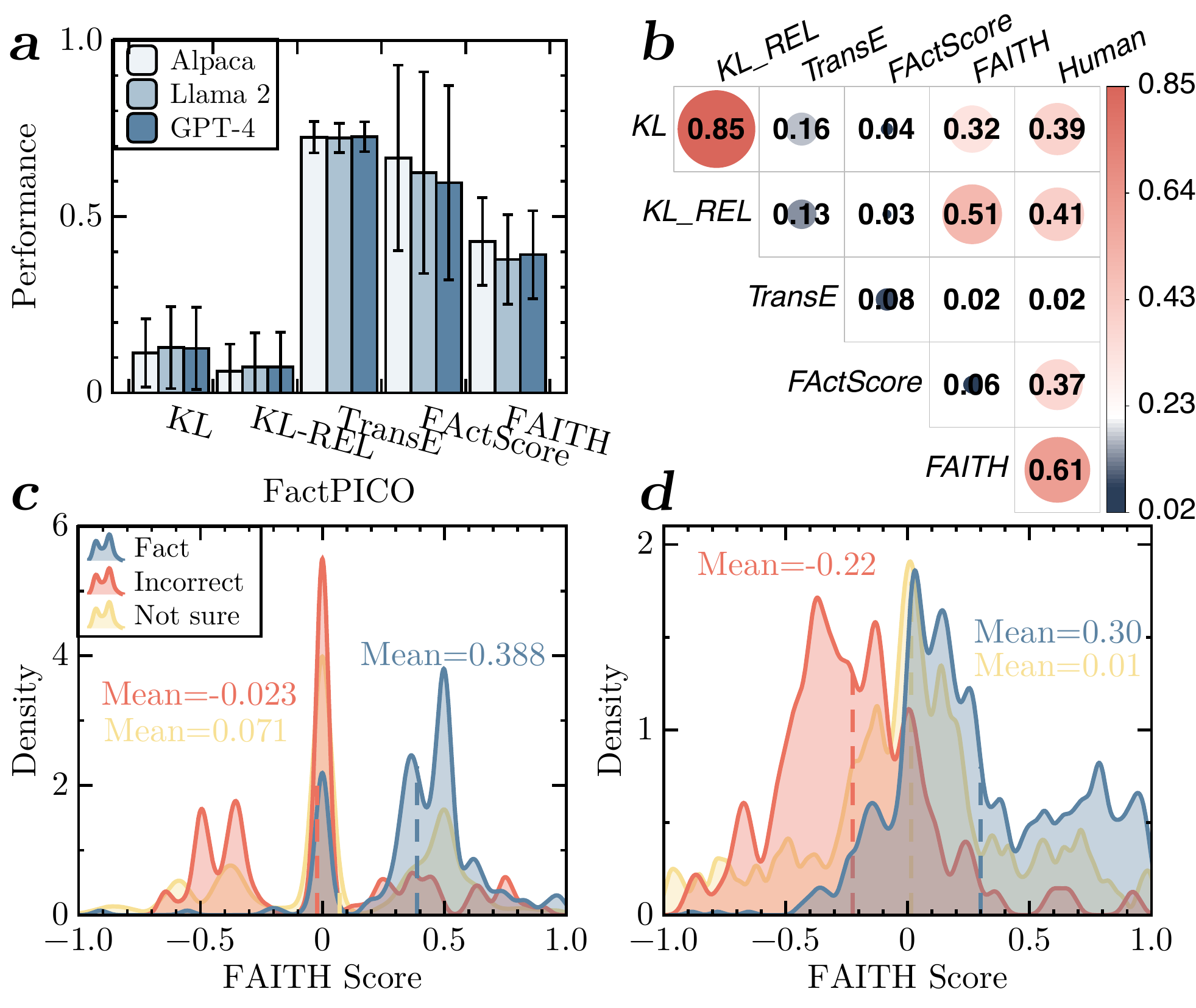}
    \caption{\small\textbf{Broad applicability of \our\ to medical summarization and fact verification (MFV). a,} Factuality scores assigned by \our\ to LLM-generated medical summaries from the FactPICO benchmark, reflecting alignment with model performance hierarchies. \textbf{b,} Pearson correlations of \our\ and baselines with expert judgements on FactPICO. \textbf{c, d,} Distribution of \our\ scores for claims of known veracity from the HealthFC (c) and BEAR-FACT (d) MFV benchmarks, illustrating clear differentiation between true and false statements.}
    \label{fig:summarization}
\end{figure}

\subsection{Broader Applicability of FAITH}
\label{sec:rq4}

To interrogate \our's utility beyond medical QA, we further assessed its performance on medical summarization~\cite{ShaibLJMLW23,TangSINSEXDDRWP23, ThawakarSMCAKLK24, VanVBLJ24} and medical fact verification (MFV)~\cite{VladikaSM24,MohrWK22} (details available in Appendix C.4). For medical summarization, we used the FactPICO benchmark~\cite{JosephCTGCXWL24}, which comprises expert factuality assessments of $345$ LLM-generated summaries of $115$ randomized controlled trials (RCTs) across the Alpaca, Llama-2 and GPT-4 models. Results shown in~\pref{fig:summarization}a indicate that \our\ aligns closely with the expert-rated model rankings (Alpaca $>$ GPT-4 $>$ Llama-2). More importantly, \our\ achieves the highest Pearson's correlation with expert judgments ($\rho=0.61$; \pref{fig:summarization}b). For MFV task, we assessed \our\ on the HealthFC~\cite{VladikaSM24} and BEAR-FACT~\cite{WuhrlRGK24} datasets, which consist of $750$ and $300$ medical claims, respectively. The results show that \our\ effectively distinguished true from false claims (\pref{fig:summarization}c, d, Fig. A17, and Fig. A18). 

\section{Ablation Studies}
\label{sec:ablation}

\subsection{Sensitivity to KG Choice and Integrity}
\label{sec:ablation_kg}
A critical question for any KG-based approach is how its effectiveness depends on the choice of KG.

\noindent \textbf{Compatibility with different KGs.} To validate this, we integrated \our\ with other established KGs of varying scales, PrimeKG~\cite{ChandakHZ23} and OGBL-biokg~\cite{HuFZDRLCL20} (Appendix B.3), and repeated our core experiments. The results show that while relative ranking of LLMs remains consistent, the performance will be affected by different KGs (Fig. A11).

\noindent \textbf{Robustness against KG noise.} Given the real-world KGs inevitably contain noise (e.g., missing/erroneous triplets), we assessed \our's robustness by perturbing the UMLS. We applied $3$ types of significant~\cite{AlbertJB00} noise: $i)$ $20\%$ random edge deletion, $ii)$ $20\%$ random node deletion (including incident edges), and $iii)$ $20\%$ random noisy edge insertion. The findings in Fig. A12 showed that deleting entire nodes or inserting noisy edges cause performance degradation. This demonstrates that high-quality KGs remain a crucial prerequisite.

\subsection{Reliability of Medical Claim Extraction Module}
\label{sec:ablation_claim}
To assess the medical claim extraction pipeline in \our, we compare various extractor implementations against QuickUMLS~\cite{SoldainiG16} on the BioRED~\cite{LuoLWAL22} and DDI13~\cite{Segura-BedmarMH13} datasets (details in Appendix C.7).

\noindent \textbf{Impact of prompting strategies.} We first tested how our multi-round and critical analysis prompting affect extraction performance. We thus implemented the LLM extractor in \our\ with $4$ prompting strategies: $\blacktriangleright$ \textit{Base prompt}: a basic extraction prompt without 
advanced techniques. $\blacktriangleright$ \textit{Base + critical}: the base prompt with critical analysis. $\blacktriangleright$ \textit{Base + multi}: the base prompt with multi-round conversation. $\blacktriangleright$ \textit{FAITH}: \our's full prompting strategy. As shown in Fig. A13, all LLM-based variants surpass QuickUMLS on both datasets. Multi-round conversation mainly boosts recall, while the critical analysis prompts helps ensure the precision, making their combination most effective.


\noindent \textbf{Sensitivity to LLM choice.} We then examine the sensitivity of the claim extraction performance to the specific choice of LLM. For this, we replaced the GPT-4o model with $4$ other LLMs and repeated the experiments on BioRED and DDI13. The results in Fig. A14 showed that while changing to cheap open-source models like Llama 3 and Llama 3.1 leads to a slight decrease, all LLM extractors still yield considerable improvements over QuickUMLS. On the other hand, the performance gain of GPT-4o over GPT-4o-mini is marginal, so GPT-4o model is already well-suited for the task.
\section{Discussion and Conclusion}
\label{sec:discussion}
This paper investigated the reliability of using medical KGs as a foundation for evaluating LLM factuality in healthcare. Through our framework, \our, we demonstrated that leveraging KGs is a powerful and reliable approach for LLM factuality evaluation in healthcare.

However, our work also illuminates the some limitations of this approach. The first one is its reliance on the KG's quality and coverage. Factual claims involving knowledge absent from the KG cannot be verified, which can lead to potential false negatives. Secondly, the entire pipeline is dependent on the performance of the upstream medical claim extraction module. While our LLM-based extractor proved effective, errors at this stage can propagate through the system. Future work should focus on mitigating these limitations, perhaps by integrating multiple KGs or developing methods to handle claims that are out-of-scope for the KG.

\noindent\textbf{Social impact.}  \our\ enhance the safety and reliability of LLMs~\cite{ZhouHS024, ZhouLM22, Zhou0M22} by automatically assessing the factuality of their responses. It serves as a crucial ``guardrail'' against misinformation, fostering trustworthy adoption by clinicians and regulators while guiding developers with targeted feedback for model improvement. The core approach is also adaptable for fact-checking in other critical domains like law~\cite{CuiLY23, HamdaniBMHS24} and finance~\cite{XieHZ+23, KangL23} where factuality is paramount.

\noindent\textbf{Code availability.} Source code of \our\ can be found at: https://github.com/COLA-Laboratory/FAITH.

\section{Acknowledgements}
We sincerely thank all the reviewers for their encouraging and constructive feedback. This work was supported by the UKRI Future Leaders Fellowship under Grant MR/S017062/1 and MR/X011135/1; in part by NSFC under Grant 62376056 and 62076056; in part by the Royal Society Faraday Discovery Fellowship (FDF/S2/251014), BBSRC Transformative Research Technologies (UKRI1875), Royal Society International Exchanges Award (IES/R3/243136), Kan Tong Po Fellowship (KTP/R1/231017); and the Amazon Research Award and Alan Turing Fellowship.
\bibliography{paper.bib}




\newpage

\end{document}